# ADDRESSING AMBIGUITY IN MULTI-TARGET TRACKING BY HIERARCHICAL STRATEGY

*Ali Taalimi, Liu Liu and Hairong Qi*

Department of Electrical Engineering and Computer Science
University of Tennessee, Knoxville, TN, USA 37996
{ataalimi, lliu25, hqi}@utk.edu

## ABSTRACT

This paper presents a novel hierarchical approach for the simultaneous tracking of multiple targets in a video. We use a network flow approach to link detections in low-level and tracklets in high-level. At each step of the hierarchy, the confidence of candidates is measured by using a new scoring system, ConfRank, that considers the quality and the quantity of its neighborhood. The output of the first stage is a collection of safe tracklets and unlinked high-confidence detections. For each individual detection, we determine if it belongs to an existing or is a new tracklet. We show the effect of our framework to recover missed detections and reduce switch identity. The proposed tracker is referred to as TVOD for multi-target tracking using the visual tracker and generic object detector. We achieve competitive results with lower identity switches on several datasets comparing to state-of-the-art.

## 1. INTRODUCTION

**Motivation:** Multi-target tracking is essential to many applications such as surveillance, robotics, and driver assistance [1, 2, 3]. The target should be detected in each frame, and correspondences between the identities of the target are drawn from frame to frame. Due to significant improvement in human detectors [4, 5, 6, 7], many recent tracking methods adhere to the tracking-by-detection (TBD) strategy, applies an offline trained object detection independently to every frame to estimate target locations throughout the video. An affinity model (defined by integrating several cues like motion and appearance) is used to link detections across frames in an association optimization framework *i.e.* "data association" task. Hence, tracking problem is converted to a data association and is reformulated as finding the optimal assignment.

**Challenges:** Although tracking-by-detection obtains state-of-the-art results, it is largely based on detectors which are not perfect. Detectors are reliable as long as targets have sufficient distance from each other. They fail when objects overlap each other objects or by fixed objects in the scene like trees and poles which leads to miss-detections, false positives, and incorrect responses. Several recent algorithms address this issue in a combinatorial optimization framework [1, 8, 9].

Early studies applied greedy bipartite data association to solve a series of bi-partite assignment problems on a frame-by-frame basis [10]. In this case, motion model will only account the distance that object is moved between two frames. which is limited and can cause switch id and drift. One effective way to solve this issue is to consider a batch of several frames rather than inferring the current state from just the previous observation of a track [11]. Using more frames leads to a more reliable affinity model as the extra frames may have information about a target that is occluded in other frames.

**Hierarchical Scheme:** Hierarchical data association [1] links detections into progressively longer track fragments (tracklets) and consequently, enables tracking-by-detection to deal with short and long time occlusions [3]. The process operates on multiple levels. At the first level, short but reliable fragments of the tracks, or "safe tracklets", are extracted. In fact the safe tracklets are high-confidence partial trajectories that assist tracking method to face with potential ambiguities. They gradually linked to each other to make long tracks.

It is evident that short-time and long-time motion characteristics of objects are different. It can be safely assumed that objects in high frame rate videos move with constant velocity in short temporal window, but, this assumption does not hold for motion in long term. Unlike methods that extract trajectories in one pass [8, 12], hierarchical data association can have different motion models for different levels of merging. Motion cue in early steps of hierarchy is linear motion model [13, 14] and in higher-level, temporal smoothness is used to penalized nonsmooth trajectories. Appearance information is accumulative in the hierarchical scheme, which facilitates finding targets in the presence of occlusion or miss detection. In addition in each step there is more appearance information is available rather previous stage, which makes it possible to find targets in complex situations like when target is occluded or is missed.

**Challenges in Hierarchical:** Extracting tracks in the hierarchical fashion strongly depends on quality and quantity of the safe tracklets. The tracklets should be reliable, meaning that they should have bounding boxes with only one target inside and that target should be clearly distinct from the scene background.

The outcome of low-level association would be a collection of tracklets and unlinked, single detections. In this step, one should determine if single detections that were not associated in the first stage, should be kept when either belong to one of the already found tracklets, or they are initiating new trackers. Otherwise, they should be discarded as spurious. Since they are single observations, the motion cue only can describe how far target moves. But at least two nodes are required to determine velocity, so it is not applicable to single detections. Some researchers assign a fixed, arbitrary cost to each single detection [15], others assign a fixed speed in the direction perpendicular to the closest image border [13]. To best of my knowledge these issues have not been addressed in literature.

**Low-Level Contribution:** We notice that most of the hierarchical trackers never address the issue of producing tracklets safely. They either start with safe tracklets or they assume that safe tracklets can be obtained based on two metrics, physical overlap of detections and detector confidence scores. But, a scene with multiple targets nearby can cause misleading detector confidence scores and large amounts of overlap between different targets. Therefore, neither of these two metrics is reliable. We proposed CanfRank [16] to assign more accurate confidence scores and distinguish true positives from false positives while keeping missed detection low. More details and its extension to measure confidence for tracklets is presented in Section 2.1.

**High-Level Contribution:** According to [17], motion cue, particularly a constant velocity constraint is the most effective way to associate tracklets. Especially in a crowded area which people may wear similar clothing and appearance information is not distinctive, motion can be vital for tracking. In low-level merging step most restrictive constraints are adopted in order to find small but accurate tracklets that cause many short tracklets and a lot of high confident single detections which are not associated with other observations. Recently, single-target trackers are fast and robust to occlusion [18]. We propose to convert each single detection to a tracklet by extending it forward or backward in time using these new single-target tracking algorithms. We elaborate our approach in Section 2.2. Finally, at the high-level of association, the goal is to recover whole tracks from the previously found tracklets. The flowchart of the proposed TVOD is illustrated in Figure 1.

## 2. PROBLEM FORMULATION

We follow the tracking-by-detection strategy and deduce people trajectories by looking at people as independent sets of hypotheses $D$ obtained by a part-based human detec-tor [19] in a preprocessing step. Let $M$ be the number of frames in a video; we denote the detection $i$ at the frame $t$ as $d_i^t = \{p_i^t, s_i^t, c_i^t\}$ where $p_i^t, s_i^t, c_i^t$ represent the position, size, and confidence of the detection, respectively. $D = \{d_i^t \mid t \in \{1, \cdots, M\}\}$ denotes the set of detections of all $M$ frames. We define a tracklet $T_i$ of the object $i$ as an ordered list of object observations starting from frame $t_i^s$ to the frame $t_i^e$ denoted as $T_i = \{d_i^j | 1 \leq t_i^s \leq j \leq t_i^e \leq t\}$. Extracting the set of $N$ human trajectories $\mathcal{T} = \{T_1, T_2, \cdots, T_N\}$ from raw observations $D$ can be formulated by MAP $\mathcal{T}^* = \arg\max_\mathcal{T} P(D|\mathcal{T})P(\mathcal{T})$. Here the likelihood $P(D|\mathcal{T})$ measures how well the extracted trajectories are successful in explaining the observations. The prior over trajectories $P(\mathcal{T})$ has the information about tracks in general in order to give more weight to tracks which meet expected characteristics of a good solution and can be described in non-Bayesian terms using log of prior as regularizer. Usually prior describes independent tracks using Markov assumption: $P(\mathcal{T}) = \prod_i P(T_i)$.

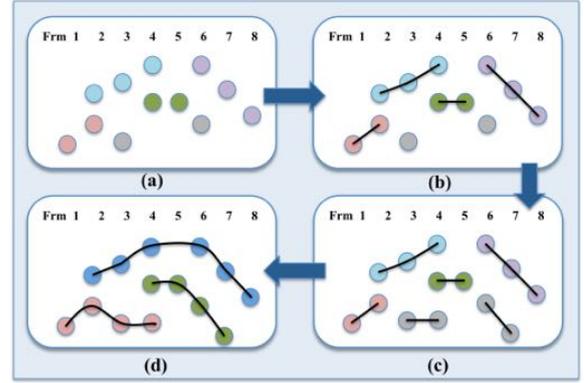

**Fig. 1**. Flowchart of the our approach. Observations obtained by detector. ConfRank is calculated for each detection based on its spatio-temporal neighborhood and confidence (a). The initial tracklets are produced by matching detections using ConfRank scores and overlap (b). The remained single detections are elongated using single-target-tracking method (c). At high-level tracklets are matched to recover whole track, while their confidence is measured by ConfRank (d).

Note that because of numerous possible combinations of $\mathcal{T}$ and $D$ it is not practical to obtain $\mathcal{T}^*$ directly. In order to propose a feasible solution to the problem, we redefine it by introducing a new method that measures confidence for detections and tracklets. Linking between candidates in each step (detections in low-level and tracklets in high-level) is modeled using a graph which all the candidates constitute the nodes and edges depict each pair to associate and their weights describe the cost of merging. In contrast to many methods which limit edges to be only between consecutive frames [8, 12], we allow edges to connect detections or tracklets with large frame gap, which let our model be capable of handling long occlusion or several frames miss detection. We consider constant velocity motion model in early stages of hierarchy, and it is changed to temporally smoothness for associating long tracklets to each other.

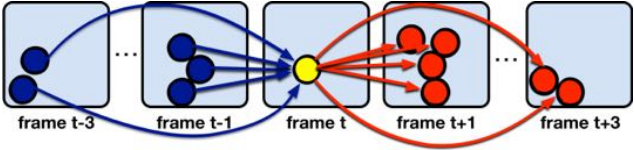

**Fig. 2**. ConfRank illustration. In the frame $t$ the yellow node $d_i^t$ receives requests from adjacent detections in previous frames (blue arrows). The yellow node is a candidate to merge to its adjacent detections in future frames (red arrows). It is easy to extend ConfRank to measure tracklets confidence.

### 2.1. Low-Level Association

Our goal in this stage is to extract the reliable fragments of the track (tracklets), by only looking at detections of small consecutive frames and extract all easy to disambiguate and not occluded parts of trajectories. These short but safe tracklets are the basis for retrieving the whole track in the end. We adopt most conservative physical and geometrical constraints to be sure to extract only the reliable fragments of the tracks. We proposed ConfRank [16] that provides a more rigid confidence scoring system to recognize false alarms. In ConfRank, nearby confident detections will boost a detections score, and similarly, unconfident detections show a hint that the area is critical and confidence scores are decreased.

The original confidence of each particular tracklet is calculated based on TC_ODAL [15]. Then, the final confidence score of each tracklet is obtained by applying ConfRank, which looks at quantity and quality of other tracklets to merge into and tracklets starting afterward which are eligible to account. In this way, we can distinguish better safe tracklets from false positives. Figure 2 illustrates that ConfRank scores each candidate (single detection or tracklets) by getting feedback from the set of candidates in spatio-temporal neighborhood of the candidate $d_i^t$.

### 2.2. Mid-Level Association - Single Detections

In this level we have safe tracklets and some high-confident single detections that are not assigned to any tracklets. Most of researches solve a bipartite matching problem for each single detection to find its associate between tracklets while the motion cost is considered as distance between single nodes and tracklets [15]. We take the inverse approach and try to elongate each single detection to be a tracklet which allows us to use constant motion model. The inverse approach is also beneficial in recovering missed detections. For each 4-frame segment the non-associated single detections with high confidence, which may arise from short occlusion, are kept as tracklets of size one. These single detections should be associated with one of the already-found tracklets or begin a new tracker. Assume $\mathbb{T}$ as first batch of tracklets so far $\mathbb{T} = \{T_i\}_{i=1}^K$ and set of isolated detections as $\Upsilon = \{\acute{d}_i | \acute{d}_i \in D, \acute{d}_i \notin \mathbb{T}\}$. Each isolated detection has only one node, so the motion cue is restricted to account for how far the target has been displaced. Since constant velocity motion model is shown to be effective in early stages [17] we are more interested to do matching based on constant-velocity smoothness term which needs more than one node.

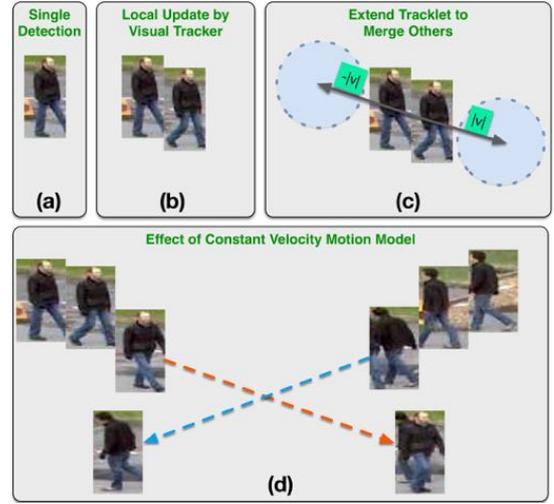

**Fig. 3**. Demonstrating the strategy for single detections. (a) single detection, (b) converting to tracklet by committee of global trackers [20] and [21] (c) predict probable locations of target, and (d) the linear motion model for short tracklets.

As Figure 3 shows, we propose to make a tracklet $\acute{T}_i$ by tracking each $\acute{d}_i^t$ back and forth within the spatio-temporal window around $\acute{d}_i^t$. We employ two different trackers and form a committee of single target trackers $\mathcal{G} = \{G_1, G_2\}$, where $G_1$ and $G_2$ are [20] and [21], respectively. We use the implementation distributed by the authors. These trackers are fast and highly robust which sounds promising for extending unlinked detections to a non-degenerated tracklet. If the decision of both trackers are close, the union of the two is assumed, otherwise, we choose the decision of the one with maximum classification score. The single target tracking algorithms need to be manually initialized by window of the target in single node.

The output of the committee $\mathcal{G}$ is a bounding box covering a part of background and foreground with no clear separation of human region from the background. This issue is a major cause behind switch id and drift, especially when multiple persons exists in one bounding box. For instance, consider the first box from left in Figure 4. It has two persons inside, and we would like tracker to associate it to the corresponding

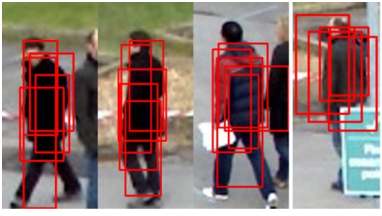

**Fig. 4**. Examples of applying interest area detector [22] on detection bounding boxes. The extracted patches mostly cover the person (foreground).

tracklet of the person who fully appear.

The generic object detector [22] highlights the foreground for tracker. We train the interest area detector [22] for pedestrian using publicly available dataset, INRIA Person [23]. Then, we run the trained detector on target window and only retain the bounding boxes that fall within 25% of the pedestrian detector [19] (Figure 4). This approach is faster and needs less information comparing to the recent trackers like [24] that rely on background subtraction either by assuming to have the background.

### 2.3. High-Level Association

The goal at this level is to extend tracklets and recover the full trajectories. We use the relationship between different tracklets to solve the association where the tracklets are modeled using a graph $G(V, E)$ with $V, E$ denoting the set of nodes and set of edges. Tracklets are nodes of the graph and edges represent the relation between tracklets and their weight reflects the probability of the tracklets belong to the same person. Without loss of generality, consider $T_j$ to appear after $T_i$ in temporal domain. The association weight of each pair $(T_i, T_j)$ is modeled using MAP formulation and will be solved using the min-cost flow algorithm. It is zero for tracklets that are not close or have frame in common. We assign edge weights as matching scores describing the probability of linking $P_{link}(T_j|T_i)$ using motion, appearance, and shape similarities based on [15].

### 3. EXPERIMENTS AND RESULTS

We evaluate our approach on 11 sequences of MOT Challenge [25]. We report tracking results based on the most widely accepted evaluation metrics, the CLEAR MOT [26] including MOTA, MOTP, MT, ML, ID switch, and fragment. We validate the claim that tracking by ConfRank instead of detector confidence effectively decreases IDS by compare its performance with the state-of-the-art hierarchical trackers TC_ODAL [15] and RMOT [27], plus a tracking by Markov Decision Processes, MDP [28] and a tracking by superpixel, SegTrack [11]. To observe the effect of each component of the proposed method, the result of three systems are reported:

**Table 1**. Comparison on MOT Challenge [25].

| Method | MOTA | MOTP | MT% | ML% | Frag | IDS |
|---|---|---|---|---|---|---|
| **RMOT**[27] | 18.6 | 69.6 | 5.3 | 53.3 | 32.0 | 17.1 |
| **TC_ODAL**[15] | 15.1 | 70.5 | 3.2 | 55.8 | 46.0 | 17.1 |
| **MDP**[28] | 30.3 | 71.3 | 13.0 | 38.4 | 31.8 | 14.4 |
| **SegTrack**[11] | 22.5 | 71.7 | 5.8 | 63.9 | 20.2 | 19.1 |
| **CR-RMOT** | 21.1 | 69.8 | 6.6 | 48.5 | 28.8 | 12.9 |
| **CR-ODAL** | 19.8 | 71.2 | 5.5 | 45.6 | 43.4 | 13.8 |
| **CRnop** | 22.3 | 70.8 | 9.4 | 43.8 | 30.2 | 14.5 |
| **TVOD** | 33.9 | 71.4 | 12.5 | 37.5 | 33.2 | 12.1 |

**CR-RMOT:** RMOT[27] with ConfRank;
**CR-ODAL:** TC_ODAL[15] with ConfRank;
**CRnop:** TVOD with visual tracker without detector [22];

Comparing CR-RMOT and CR-ODAL with RMOT and TC_ODAL shows that when measured by MOTA, the most comprehensive metric for tracking performance, tracking by ConfRank consistently yields superior performance as compared to tracking by detector confidence for more than 3.5%. Also, IDS and Frag is reduced by more than 3.3% and 2.6%. Also, TVOD effectively reduced IDS and increases MOTA while shows a reasonable performance comparing with MDP and SegTrack. While MOTA and precision have obvious improvements for TVOD, MT and ML are enhanced and other tests are competitive. Since in low-level we applied ConfRank to make safe tracklets, we have less false positives and switch identities with the cost of increasing fragmentation. The processing time for TC_ODAL, RMOT and TVOD is 0.19, 0.21 and 0.23 seconds per frame, respectively.

### 4. CONCLUSION

In this paper, we proposed a new method for multi-target tracking in hierarchical scheme where candidates are detections in low-level and tracklets in high-level linked in multiple levels. Specifically, we studied two important challenges in hierarchical tracking, how to find safe candidate (detection and tracklet), and how to deal with unlinked but high-confident detections. We update the confidence of each candidate by looking at its spatio-temporal neighborhood. The intuition is that a highly confident detection in one frame can vote for nearby detections in neighboring frames, and helps to improve the confidence score in ambiguous situations. Also, unconfident detection (like links from false detection) decrease nearby detection scores in neighbouring frames. Also, we introduce a simple but effective local updates to resolve ambiguous situations by extending unlinked detections to become a tracklet. This is done using combination of single target trackers and interest area detector.